\documentclass[10pt,twocolumn,letterpaper]{article}
\usepackage[pagenumbers]{cvpr}
\usepackage{graphicx}
\usepackage{amsmath}
\usepackage{amssymb}
\usepackage{booktabs}
\usepackage{algorithm}
\usepackage{algorithmic}
\usepackage{amsfonts} 
\usepackage{bbm}
\usepackage{bm}
\usepackage{amsthm}
\usepackage{mathrsfs}
\usepackage{indentfirst}
\usepackage{multirow}

\usepackage[pagebackref,breaklinks,colorlinks]{hyperref}

\usepackage[capitalize]{cleveref}
\crefname{section}{Sec.}{Secs.}
\Crefname{section}{Section}{Sections}
\Crefname{table}{Table}{Tables}
\crefname{table}{Tab.}{Tabs.}

\begin{document}

\title{Trimap-guided Feature Mining and Fusion Network for  Natural Image Matting}
\author{Weihao Jiang$^{1}$, Dongdong Yu$^{2}$, Zhaozhi Xie$^{1}$, Yaoyi Li$^{1}$, Zehuan Yuan$^{*,2}$, Hongtao Lu$^{*,1}$\\
$^{1}$Shanghai Jiao Tong University, China \quad 
$^{2}$ByteDance Inc.\\
{\tt\small \{jiangweihao, xiezhzh, dsamuel, htlu\}@sjtu.edu.cn} \\
{\tt\small \{yudongdong,  yuanzehuan\}@bytedance.com}
}
\maketitle
\let\thefootnote\relax\footnotetext{$^{*}$ Z. Yuan and H. Lu are the corresponding authors.}
\let\thefootnote\relax\footnotetext{This work was performed while Weihao Jiang worked as an intern at ByteDance.}
\begin{abstract}
   Utilizing trimap guidance and fusing multi-level features are two important issues for trimap-based matting with pixel-level prediction. To utilize trimap guidance, most existing approaches simply concatenate trimaps and images together to feed a deep network or apply an extra network to extract more trimap guidance, which meets the conflict between efficiency and effectiveness. For emerging content-based feature fusion, most existing matting methods only focus on local features which lack the guidance of a global feature with strong semantic information related to the interesting object. In this paper, we propose a trimap-guided feature mining and fusion network consisting of our trimap-guided non-background multi-scale pooling (TMP) module and global-local context-aware fusion (GLF) modules. Considering that trimap provides strong semantic guidance, our TMP module focuses effective feature mining on interesting objects under the guidance of trimap without extra parameters. Furthermore, our GLF modules use global semantic information of interesting objects mined by our TMP module to guide an effective global-local context-aware multi-level feature fusion. In addition, we build a common interesting object matting (CIOM) dataset to advance high-quality image matting. Particularly,
results on the Composition-1k and our CIOM show that our
TMFNet achieves 13\% and 25\% relative improvement on SAD,
respectively, against a strong baseline with fewer parameters
and 14\% fewer FLOPs. Experimental results on the Composition-1k test set, Alphamatting benchmark, and our CIOM test set demonstrate that our method outperforms state-of-the-art approaches. 
\end{abstract}

\section{Introduction}
\label{sec:intro}

As a kind of image matting task, the alpha matting task separates foreground objects in images by predicting an alpha matte which represents the opacity of the foreground at each pixel. In a mathematical form, alpha matting defines the natural image $I$ as a convex
combination of a foreground image $F$ and a background image $B$ at each pixel $i$, as shown below:
\begin{equation}
	I_i ={\alpha}_i F_i + (1-{\alpha}_i) B_i, {\alpha}_i \in [0,1],
	\label{alphaeq}
\end{equation}
\begin{figure}[t]
		\centering
		\includegraphics[width=0.98\linewidth]{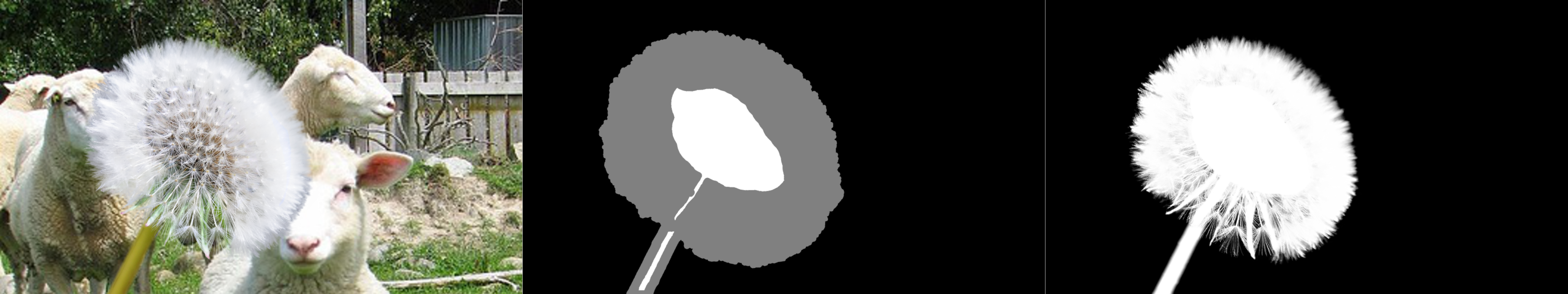}
		\caption{ The illustration of trimap-based matting. Columns from left to right, an input image, a trimap, and a ground-truth alpha matte, respectively. The black region, white region, and gray region in trimap denote background, foreground, and unknown regions, respectively. The non-background area points out an interesting object.}
		\label{trimap}
\end{figure}
where ${\alpha}_i$ is the value of the predicted alpha matte at pixel $i$. As is explained in \cite{gca,adamatting}, the problem is highly ill-posed. To solve this problem, many approaches \cite{bayes_12,iterative} utilize trimap or scribble as constraint information to reduce the solution space. Trimap-based approaches can not only reduce the solution space by using trimap, but also know which object should be treated as the foreground object  in a complex image based on the guidance of non-background area in input trimap. As is shown in Fig.~\ref{trimap}, the input trimap points out which objects should be predicted as a foreground for both overlapped objects and objects close to each other. Since other salient but not interesting objects such as parts of sheep can also appear in the unknown regions of trimap in Fig.~\ref{trimap}, a good trimap-based approach should utilize the semantic guidance of input trimap to predict the opacity of interesting object, instead of simply predicting all salient objects in unknown regions.

Deep learning methods have  achieved significant improvements in trimap-based matting tasks in recent years. Most of them \cite{deepmatting,indexnet,context,fba,pii,lfpnet} utilize trimap information  by directly concatenating input image and input trimap to feed an encoder network. Some of them \cite{adamatting,sim,timinet} learn or process trimap information with  an extra network. Attention-based methods such as GCA \cite{gca} and HDMatt \cite{hdmatt} propagate the information flow between different regions indicated by the trimaps, depending on the similarity between patches of keys and patches of queries. However, these approaches neglect the strong high-level semantic cues for interesting objects provided by non-background regions in input trimaps and have not utilized them to mine high-level semantic information of interesting objects in an efficient way.

Fusing or aligning the low-resolution high-level features and high-resolution low-level features is another important issue for image matting. Most approaches \cite{deepmatting,gca,fba,sim}
adopt static methods which upsample high-level features by transposed convolutions or bilinear upsampling and then fuse them with low-level features by addition or a convolution layer.   Advanced content-based methods emerge in recent years.  IndexNet \cite{indexnet}, CARAFE \cite{carafe} and A$^2$U \cite{a2u} adopt content-based spatial dynamic upsampling by predicting content-aware upsampling kernel instead of distance-based upsampling or static transposed convolution. Considering that semitransparent parts of a given foreground object may have different appearances in different background scenes, spatial dynamic fusion kernels may work better than  static convolution kernels.  However, these content-based approaches fuse the high-level and low-level features only depending on local features, which may neglect the global context feature with high-level semantic information closely related to the interesting objects.

In this paper, we propose a trimap-guided feature mining and fusion network (TMFNet) which mines high-level semantic information of the interesting object under trimap guidance efficiently and fuse multi-level features with global-local context-aware spatial dynamic kernels effectively. The proposed TMFNet mainly consists of our trimap-guided non-background multi-scale pooling (TMP) module and  global-local context-aware feature fusion (GLF) modules. 

We propose TMP module to mine semantic context information of interesting objects by utilizing the high-level semantic guidance of input trimap without extra parameters. Global pooling and large kernel pooling such as the pyramid pooling module \cite{psp} are widely used in both image matting \cite{fba} and semantic segmentation \cite{upernet}  to capture semantic information from the global context. As shown in Fig.~\ref{trimap}, trimap-based matting needs to separate interesting objects pointed out by non-background regions in trimap  instead of every object, which inspires us to aggregate the weighted average from the non-background area of high-level semantic features in different scales. Since image matting requires more spatially accurate prediction, we combine non-background average pooling and  multi-scale pooling kernels with a stride of 1 to build our TMP module, which integrates the high-level semantic cues of trimap into the semantic context information mining of interesting objects without extra parameters.

Different from previous content-based aligning or fusion methods \cite{indexnet,a2u,carafe}, our GLF modules not only utilize local features, but also introduce the global context feature mined from our TMP module to predict better dynamic fusion kernels efficiently and effectively. Since both global context and local feature are important for pixel-level prediction, the proper selection of a global feature improves our feature fusion significantly.

In addition, we build a common interesting object matting  dataset to advance high-quality and high-resolution trimap-based image matting.

Our major contributions can be summarized as follows:
\begin{itemize}
\item We propose a trimap-guided non-background multi-scale pooling (TMP) module to mine semantic information of interesting objects, which utilizes high-level semantic guidance in trimap without extra parameters.

\item We design a novel lightweight global-local context-aware feature fusion (GLF) module which introduces a global feature with high-level semantic information mined from our TMP module to promote the generation of fusion kernels efficiently and effectively.

\item  We build a common interesting object matting (CIOM) dataset to advance high-quality and high-resolution trimap-based natural image matting.
\item Experimental results on Composition-1k \cite{deepmatting} test set, Alphamatting \cite{alphamatting} benchmark, and our CIOM test set demonstrate that our TMFNet outperforms the state-of-the-art approaches in natural image matting.
\end{itemize}

\section{Related Work}

\textbf{Trimap-based image matting.} In general, trimap-based image matting methods fall into three main categories: sampling-based methods, affinity-based methods, and learning-based methods. Sampling-based methods \cite{bayes_12,cluster_14,global_18,alpha_36} solve Eq.~\ref{alphaeq} by sampling colors from background and foreground regions for each pixel in the unknown region. Propagation methods \cite{knn_10,random_17,closed_24,spectral_25} estimate alpha by propagating its values from known regions to unknown regions based on the color line model proposed by \cite{closed_24}. Benefiting from the development of powerful deep convolution networks and a large-scale matting dataset \cite{deepmatting}, most trimap-based approaches \cite{deepmatting,indexnet,context,fba,pii,lfpnet} utilize the semantic cues of input trimap by concatenating images and trimap to feed the network. Some approaches use an extract network to learn or extra semantic information from input trimap. ADA \cite{adamatting} uses an extra decoder to learn an adapted trimap and propagates it with the output of the alpha decoder. SIM \cite{sim} uses an extra patch-based classifier to generate their semantic trimap, which needs their extended dataset with class labels of foreground objects. TIMINet \cite{timinet} uses an extra network component to mine trimap information. Non-local matting approaches \cite{gca,hdmatt}  guide information flow from the image context to unknown pixels in trimap using an attention mechanism.

\textbf{Fusing or aligning high-level features and low-level features.} CNN-based models usually provide high-level features with low resolution and low-level features with high resolution. Therefore, it is an important issue to fuse or align high-level and low-level features for deep matting tasks. Existing approaches fall into two main categories: static method and content-based method. Most approaches \cite{deepmatting,fba,gca,lfpnet} adopt static methods, which upsample high-level features with bilinear kernel or transposed convolution and the upsampled high-level features fused with low-level features by convolution or direct addition. Other approaches \cite{indexnet,carafe,a2u} adopt content-based spatial dynamic upsampling  instead of static upsampling to advance feature fusion. IndexNet \cite{indexnet} and CARAFE \cite{carafe} generate upsampling kernels according to  first-order features, while A$^2$U \cite{a2u} uses second-order information to generate its upsampling kernels. All these methods only use local features for their dynamic or static fusion.

\begin{figure*}[t]
  \centering
  \includegraphics[width=0.97\linewidth]{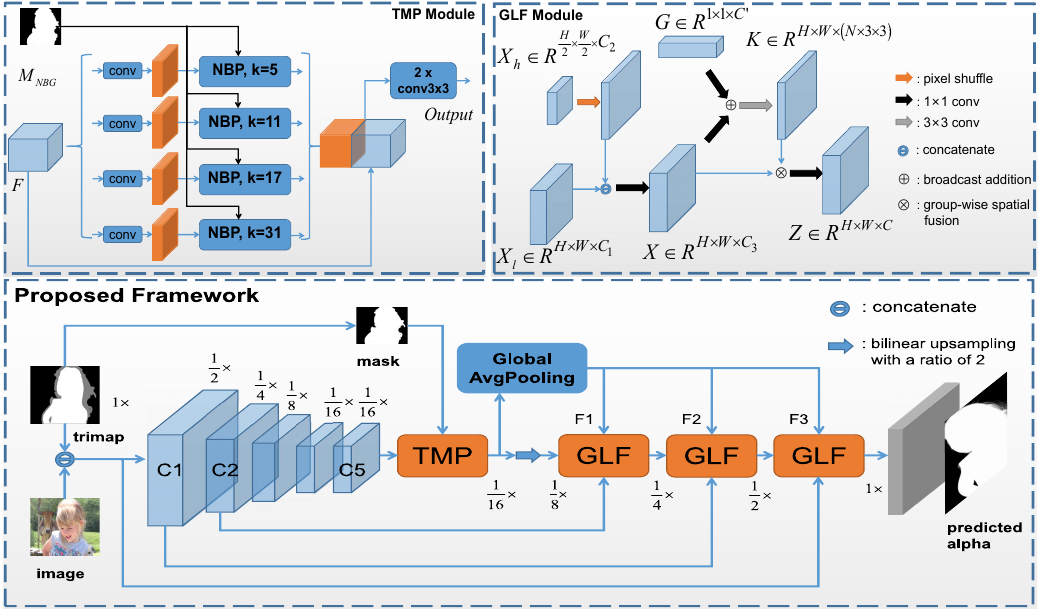}
  \caption{Our proposed TMP module, GLF module and framework of our TMFNet.}
  \label{fig:fra}
\end{figure*}

\section{Our Baseline for Deep Alpha Matting}
\textbf{Baseline Structure.} Our encoder is a ResNet-50 \cite{resnet} like \cite{sim} with an output stride of 16. An image and a one-hot trimap are concatenated as a 6-channel input like \cite{gca}. The input is fed to our encoder to generate different levels of features. Our baseline decoder firstly passes the output of C5 stage in ResNet-50 to a Pyramid Pooling layer \cite{psp} followed by two $3\times 3$ convolution layers  which is denoted as ``ppm''. Then the high-level output features of ``ppm'' will be fused with low-level features from stages of C2, C1, and the 6-channel input through a bilinear upsampling, concatenation and a convolution layer with a Leaky ReLU \cite{leaky} in sequence. We denote this fusing process in the baseline as ``static fusion'' in this paper. Finally, the fused features are fed to two convolution layers to predict the alpha matte. More details are reported in supplemental materials.

\textbf{Loss Function.} We adopt the alpha loss $L_{\alpha}$ and composition loss $L_{c}$ in DIM \cite{deepmatting} as our base loss, then a Laplacian loss $L_{lap}$  \cite{context} is added to the total loss, which is shown as:
\begin{equation}
	L_{total} = 0.5L_{\alpha} + 1.5L_{c} +0.2L_{lap},
	\label{loss}
\end{equation}

\section{A New Image Matting Dataset}
Focusing on high-quality interesting object image matting, we collect 733 high-quality images with common interesting foreground objects without motion blur. We manually  extract their alpha
mattes and foreground images with Photoshop. We select 683 labeled alpha mattes with corresponding foregrounds as our training set which are composited onto the background images from COCO \cite{coco} during training. The other 50 labeled images are composited onto background images from ADE20k \cite{ade} to form 1000 test images according to the composition rules in \cite{deepmatting}. The average number of pixels for Composition-1k  \cite{deepmatting} training images and test images are $1329^2$ and $1511^2$, respectively. The average number of pixels for our CIOM training images and test images are $2856^2$ and $2674^2$, respectively, which makes it more suitable for matting of higher resolution and quality. More details are in supplemental materials. 

\begin{figure*}[t]
	\centering
	\includegraphics[width = 0.95\textwidth]{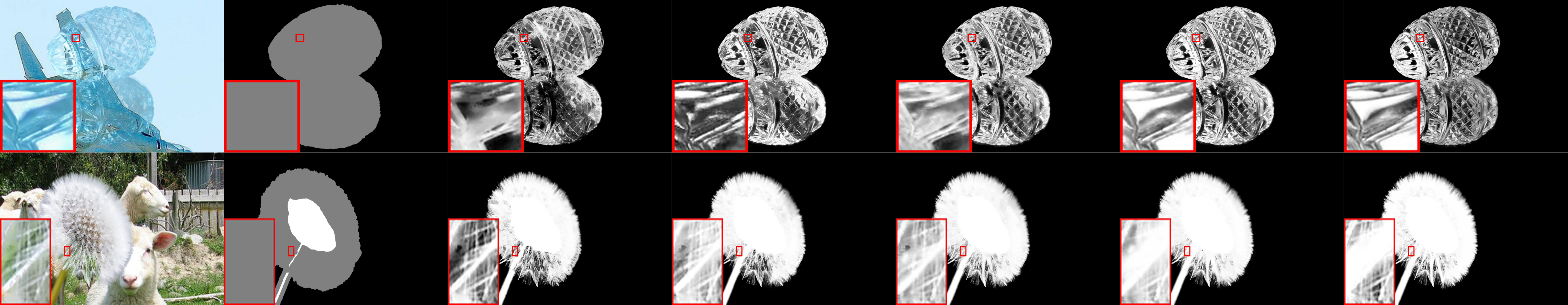}
	\caption{The visual comparison results on Composition-1k \cite{deepmatting} test set. From left to right, the original image, trimap, IndexNet \cite{indexnet}, GCA \cite{gca}, baseline, ours and ground-truth. }
	\label{fig:comp1k}
\end{figure*}

\section{Proposed Methods}
\subsection{The TMP Module}
To construct a powerful semantic representation for complex scenes for image segmentation, the pyramid pooling module \cite{psp} aggregates global context information and sub-region context with large-kernel pooling and fuses them. However, trimap-based image matting approaches need to focus on the interesting object pointed out by trimap instead of every object in context.  Since input trimap contains high-level semantic guidance, passing it with low-level images through the same network structure is not enough to extract its  semantic guidance. Observing that non-background regions of trimap are closely related to interesting objects, it is reasonable to focus the semantic feature mining on the non-background regions. Considering that image matting needs a more smooth representation, it is proper to extract a powerful and smooth  semantic representation for interesting objects by fusing the weighted average of high-level features on non-background regions with different-scale large kernels with a stride of 1. In this way, we are able to integrate the high-level semantic guidance of trimap into mining semantic information of interesting objects without extra network components.

With the above analysis, we  introduce the trimap-guided non-background multi-scale pooling module (TMP), which provides powerful semantic representation for interesting objects pointed out by trimap efficiently. Our TMP takes a high-level feature map $F$ and a non-background weighted mask $M_{NBG}$ in the same spatial size of $F$ as inputs. To get $M_{NBG}$, we generate a non-background binary mask  $\mathbbm{1}_{s \notin B}$, where $B$ is the background region in trimap, and then bilinearly resize it  in  the same spatial size of $F$ to get  $M_{NBG}$. As is shown in Fig.~\ref{fig:fra}, We reduce the channels of input feature map $F$ with 4 parallel $1\times 1$ convolution layers to get 4 reduce features denoted as $F_r$s. To focus feature mining on non-background region, we build the non-background pooling unit $NBP_k$ as:
\begin{equation}
	NBP_k(F_r,M_{NBG}) =  \frac{Pool_{k}(F_r\odot M_{NBG})}{Pool_{k}(M_{NBG})+\epsilon} ,
	\label{uint}
\end{equation}
where $Pool_k$ is an average pooling layer with a kernel size of  k and a stride of 1, $\odot$ is the Hadamard product, $F_r$, $M_{NBG}$ and $\epsilon$  are the reduced feature map, non-background weighted mask and $1e^{-6}$, respectively. Then we use 4  non-background pooling units with different kernel sizes to harvest the semantic information of interesting objects from these reduced feature maps $F_r$s. Finally, the outputs of non-background pooling units are concatenated with the high-level input feature $F$ and then they are fused by two $3\times 3$ convolution layers to form the  multi-scale context representation with high-level semantic information for the interesting object. We set the kernel sizes of non-background pooling units to 31, 17, 11, and 5 corresponding to bin sizes of 1, 2, 3, and 6 in  ``ppm'' with a $32\times 32 $ input, respectively. When the input resolution of images is $512\times 512$, our TMP has similar kernel sizes for pooling kernels  with the ``ppm''. What's more, our TMP also has the same parameter size as the ``ppm'' in  the baseline.

\subsection{The GLF Module}
Existing approaches use static methods \cite{fba,gca}  or content-based  methods \cite{carafe,a2u} to upsample a high-level feature map, then concatenate it with the low-level one, and fuse them by a convolution layer. However, these methods only focus  on  local features. 

Since both local details and global context are important for matting an interesting object, we construct our global-local  context-aware feature fusion (GLF) module utilizing local features of high-level and low-level feature maps and a global feature with high-level semantic information in a proper way. 

As is briefly shown in Fig.~\ref{fig:fra}, our GLF module firstly uses pixel shuffle \cite{ps} to align the spatial sizes of high-level feature $X_h$ and low-level feature $X_l$, and then concatenates them together. Then we use a 1$\times$1 convolution layer to distribute their information into N groups of channels in $X$. After that, N groups of $3\times 3$ kernels at each spatial position $K \in \mathbb{R}^{H \times W\times (N\times 3 \times 3)}$ are generated from both local features of $X$ and the global feature $G$. And then channels in the same group share a kernels map to fuse the spatial information. Finally,   a 1$\times$1 convolution fuses information from different groups. In this way, our GLF module fuses a high-level feature and a low-level feature under the guidance of a global feature efficiently. The exact mathematical description is as follows.

Given a low-level feature  $X_l \in \mathbb{R}^{ H \times W\times C_1}$, a high-level feature  $X_h \in \mathbb{R}^{\frac{H}{2} \times \frac{W}{2}\times C_2}$, and  a global feature $G\in \mathbb{R}^{1 \times 1\times C'}$ as inputs, the GLF module firstly distributes information of $X_l$ and $X_h$ along channel dimension as:
\begin{equation}
	X = conv_{1\times 1}(concat(PS(X_h),X_l)) ,
	\label{fuse1}
\end{equation}
where $X \in \mathbb{R}^{H \times W\times C_3}$, $conv_{1\times 1}$, $concat$ and $PS$ are internal feature map, $1\times 1$ convolution, concatenate and pixel shuffle \cite{ps}, respectively. Then we generate N groups of $3\times 3$ kernels $K\in \mathbb{R}^{H \times W\times (N\times 3 \times 3)}$ at each spatial position as:
\begin{equation}
	K = conv_{3\times 3} (Leaky(conv_{1\times 1}(X)\oplus conv_{1\times 1}(G))) ,
	\label{kernel}
\end{equation}
where $Leaky$ is the Leaky ReLU \cite{leaky} with a negative slope of 0.01 and $\oplus$ is the broadcast addition. We divide the kernels and features into N groups viewed as $K \in \mathbb{R}^{H\times W\times N \times 3\times 3}$ and $X\in  \mathbb{R}^{H\times W \times \frac{C_3}{N}\times N}$, respectively. Then a spatial fusion for each group ($\otimes$ in Fig.~\ref{fig:fra} for our GLF module) is implemented as:
\begin{equation}
	Y_{i,j,k,g} = \sum\limits_{u=-1}^{1} \sum\limits_{v=-1}^{1}K_{i,j,g,2+u,2+v}X_{i+u,j+v,k,g},
	\label{dynamic}
\end{equation}
where $g \in [1,N]$ and $k \in [1,\frac{C_3}{N}]$ are indices of groups and indices of channels in each group, respectively, and $u$ and  $v$ are offsets over $3\times 3$ kernels in $K$ at each position, respectively. Finally, $Y$ is reshaped to $Y\in  \mathbb{R}^{H \times W\times C_3}$ and we use a $1\times 1$ convolution to fuse information between groups and channels as:
\begin{equation}
	Z = Leaky(BN(conv_{1\times 1}(Y))),
	\label{fuse2}
\end{equation}
where $Z\in \mathbb{R}^{H \times W\times C}$ is the final output of GLF module and $BN$ is batch normalization \cite{bn}.

Since the high-level feature will be fused with 3 low-level features in sequence, selecting a proper global context feature $G$ is essential for our GLF Module. It is trivial to get high-level global context by applying a global pooling to $X_h$, which is denoted as GLF(B) in Table~\ref{tab:albation}. However, after $X_h$ is fused with $X_l$, the high-level semantic information is decreased and the information of local details is increased, which makes it improper to provide global context information for the next GLF module. Since our TMP module mines strong semantic information for  objects pointed out by trimap, we apply a  global average pooling to its output to generate the global feature $G$ for GLF module denoted as GLF in  Table~\ref{tab:ciom} and Table~\ref{tab:albation}. We also compare with the global feature from the input of our TMP module, namely the output of the C5 stage in ResNet-50, which is denoted as GLF(C) in Table~\ref{tab:albation}.

\subsection{Framework of TMFNet}
We replace ``ppm'' and ``static fusion'' modules in the baseline network with our TMP module and GLF modules respectively to form our proposed network. As shown in Fig.~\ref{fig:fra}, fusion stages from left to right are called F1, F2, and F3, respectively. The arrows pointing to GLFs from left, top, and bottom denote the inputs of $X_h$, $G$, and $X_l$, respectively. For GLF modules, we set their internal channel numbers $C_3$s 
to 256, 256, and 32 for stages of F1, F2, and F3, respectively and we set 16 channels for each group in all fusion stages. The output channel numbers of GLFs are the same as the baseline's. Since the spatial size of high-level feature input should be $\frac{1}{2}$ of the low-level feature's in the GLF module, we place a bilinear upsampling layer with a ratio of 2 before the F1 stage. In this way, our proposed network costs 0.9M parameters fewer and 14\% fewer FLOPs (see Tab.~\ref{tabsupp:GFLOPs}) than our baseline due to our lightweight GLF modules.

\begin{figure*}[h]
	\centering
	\includegraphics[width = 0.94\textwidth,height=4.75cm]{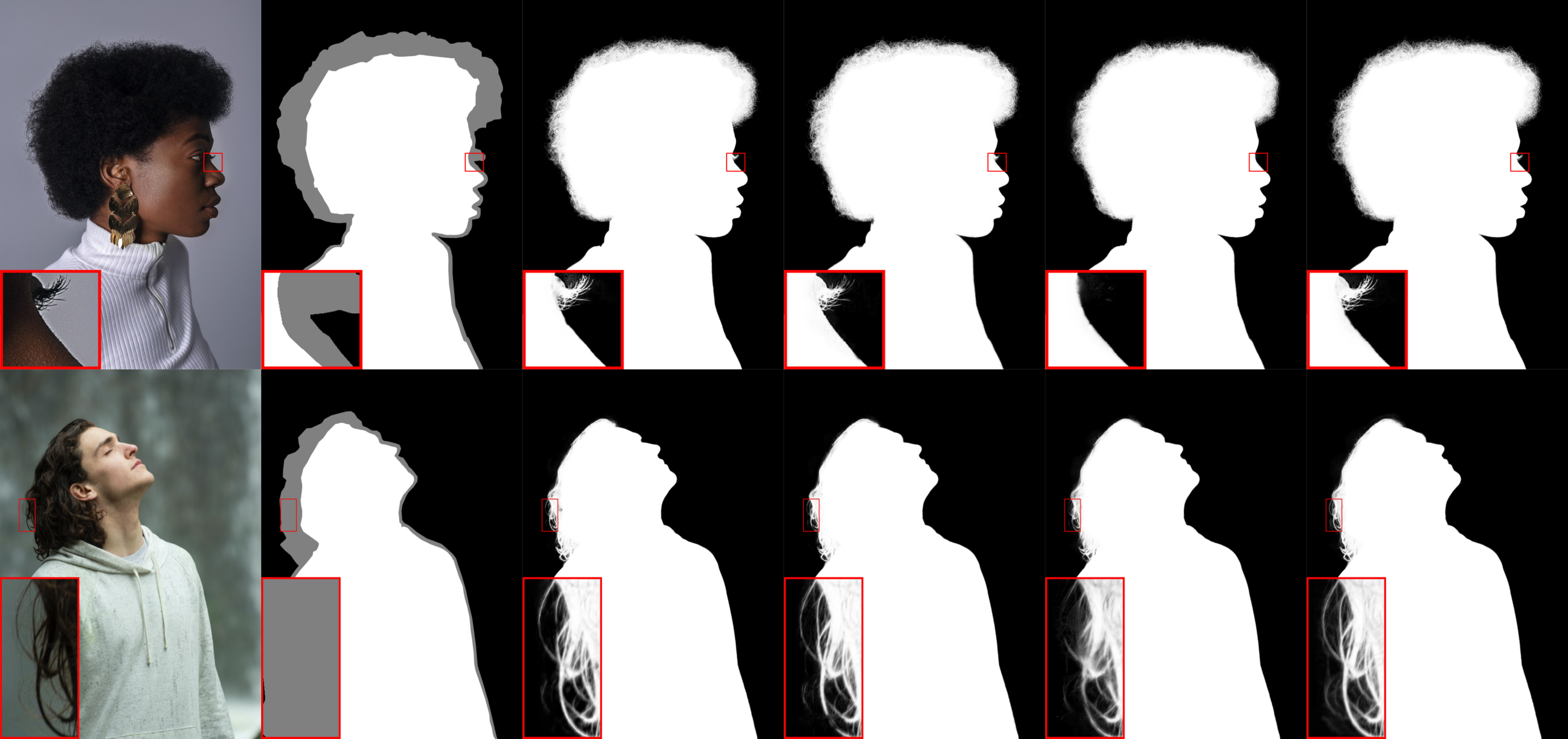}
	\caption{The visual comparison results on high-resolution real-world images. From left to right, the original image, trimap, DIM \cite{deepmatting}, IndexNet \cite{indexnet}, GCA \cite{gca} and ours. }
	\label{fig:real}
\end{figure*}

\section{Experiments}
\subsection{Experiment Settings}

Our proposed method is evaluated on Composition-1k \cite{deepmatting}, Alphamatting \cite{alphamatting} and our CIOM datasets with  quantitative results.

\textbf{Alphamatting} \cite{alphamatting} is an online real-world matting benchmark,
which provides 27 images and alpha mattes for training and 8 testing images with 3 trimaps for each one for evaluation.

\textbf{Composition-1k} \cite{deepmatting} provides 431 and 50 pairs of foreground images and alpha mattes for training and test, respectively. 1000 testing images is generated by compositing each of 50 test pairs onto 20 background images from VOC \cite{voc}, and a corresponding trimap is provided for each testing image. The backgrounds for training are from COCO dataset \cite{coco}.

\textbf{Our CIOM} provides 683 and 50 pairs of foreground images and alpha mattes for training and test. And the resolution of testing images is up to $4000\times 3867$, which can provide quantitative results for high-resolution and high-quality matting.

\textbf{Evaluation Metrics.}  We evaluate the quantitative results using metrics of the Sum of Absolute Differences (SAD),
Mean Squared Error (MSE), Gradient error (Grad) and Connectivity error (Conn).

\textbf{Implementation Details.} The baseline and proposed methods are trained for 200, 000 iterations with a batch size of 32 in total for detailed ablation study in Table~\ref{tab:ciom},~\ref{tab:albation}, and~\ref{tabsupp:glf} using 2 Tesla V100 GPUs. Especially, to compare with  SOTA methods in Table~\ref{tab:comp1k} and~\ref{tab:alphamatting}, the baseline and proposed framework are trained with a batch size of 64 in total using 4 Tesla V100 GPUs. We use Adam  optimizer \cite{adam} with an  initialized  learning rate of 0.01. The policy of learning rate decay follows GCA \cite{gca}. For data augmentation, we follow the training protocol
proposed in \cite{gca,hop}, including a random composition of two foreground images, random resizing images with random interpolation, random affine transformation and color jitters. Trimaps are generated by a dilation and an erosion on alpha images with random kernel sizes from 1 to 30 during training. The $512\times 512$ patches centered on an unknown region are cropped and  composite with a
random background image from COCO \cite{coco}. Models trained on Composition-1k \cite{deepmatting} training set or CIOM training set follow the same settings above.  As for testing, our proposed method  inferences each image without scaling in  our CIOM test set or the Composition-1k \cite{deepmatting} test set  as a whole on a single 32GB Tesla V100 GPU.
\subsection{Comparison with Prior Work}
We  compare our method with other SOTA deep trimap-based image matting methods, including LFPNet \cite{lfpnet}, FBA \cite{fba}, SIM \cite{sim}, TIMINet \cite{timinet}, GCA \cite{gca}, A$^2$U \cite{a2u}, ADA \cite{adamatting}, IndexNet \cite{indexnet} and DIM \cite{deepmatting}. 
\begin{table}[h]
    \begin{center}
     \caption{Quantitative results on the Composition-1k test set. $^{\dagger}$ denotes results with test-time augmentation. * denotes training or pre-training with extra matting data.}
    \begin{tabular}{l|cccc|c}
        \hline
        Methods & SAD & MSE  & Grad & Conn & Params \\
        \hline

        DIM\cite{deepmatting}                & 50.4 & 14.0 & 31.0 & 50.8 &\textgreater 130M\\
        Index\cite{indexnet}           & 45.8 & 13.0 & 25.9 & 43.7&8.2M \\
        ADA \cite{adamatting}        &41.7& 10.0&16.9&-&-\\
        GCA \cite{gca}        & 35.3 & 9.1 & 16.9 & 32.5 &25.3M \\
        A$^{2}$U \cite{a2u}     & 32.2 & 8.2 & 16.4 & 29.3 &8.1M \\
        TIMI \cite{timinet}    & 29.1 & 6.0 & 11.5 & 25.4 & -\\
        SIM \cite{sim}        & 28.0 & 5.8 & 10.8 & 24.8 & $\approx$ 70M\\
        FBA$^{\dagger}$ \cite{fba}        & 25.8 & 5.2 & 10.6 & 20.8 & 34.7M\\
        LFP$^{\dagger}$*\cite{lfpnet}  & 22.4 & 3.6 & 7.6 & \textbf{17.1} & 141M\\
        \hline
        Baseline & 26.4 & 4.7 & 9.3 & 22.4& 34.8M \\
        Ours & 23.0 & 4.0 & 7.5 & 18.7& 33.9M \\
        Ours $^{\dagger}$ & \textbf{22.1} & \textbf{3.6} & \textbf{6.7} & 17.6 &33.9M\\
        \hline
    \end{tabular}
    \label{tab:comp1k}
    \end{center}
\end{table}
\begin{table}[h]
    \begin{center}
    \caption{Quantitative results on our CIOM test set.}
    \begin{tabular}{l|cccc}
        \hline
        Methods & SAD & MSE  & Grad & Conn \\
        \hline
        DIM \cite{deepmatting}          & 39.7 & 6.7 & 13.4 & 35.4 \\
        Index \cite{indexnet}           & 32.5 & 5.2 & 11.4 & 28.0 \\
        GCA  \cite{gca}       & 32.1 & 6.8 & 18.2 & 25.9 \\
        \hline
        Baseline & 27.0 & 3.0 & 7.9 & 20.9 \\
        Ours(TMP) & 22.4 & 2.2 & 5.8 & 15.2 \\
       Ours(TMP+GLF) & \textbf{20.2} & \textbf{1.8} & \textbf{4.8} & \textbf{13.6} \\
        \hline
    \end{tabular}

    \label{tab:ciom}
    \end{center}
\end{table}
\begin{figure}[t]
		\centering
		\includegraphics[width=0.95\linewidth]{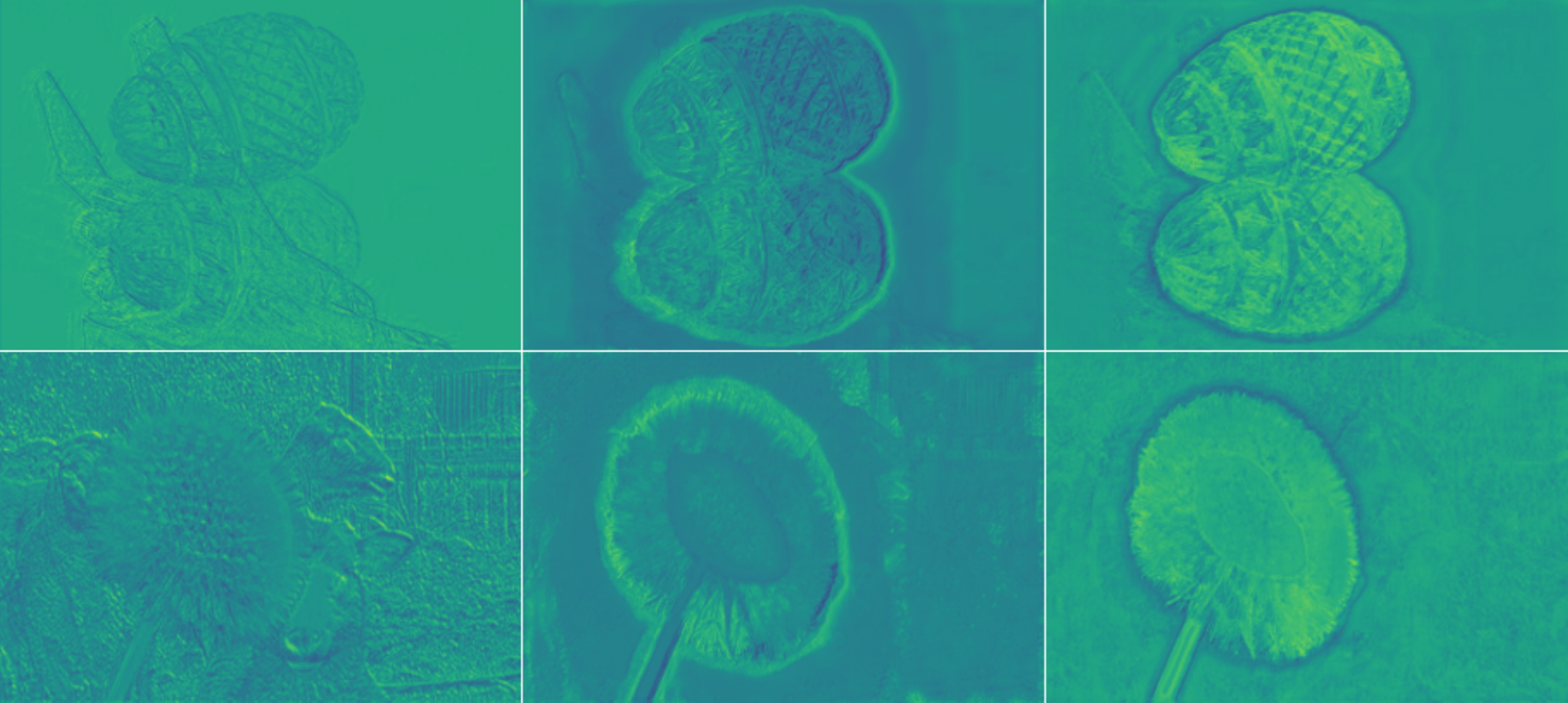}
		\caption{ Visualization of our fusion kernel maps. Columns from left to right, initial fusion kernels, kernels predicted by a ``LF'' module,   kernels predicted by a GLF module, respectively.}
		\label{fig:kernel}
\end{figure}

\textbf{Composition-1k test set.} Quantitative and visual results are reported in Table~\ref{tab:comp1k} and Fig.~\ref{fig:comp1k}. The proposed model achieves 22.1 SAD on the Composition-1k test set which outperforms other SOTA methods on the Composition-1k test set without using extra data or annotation. The proposed model also achieves significant improvements on our strong baseline model using fewer parameters. As is shown in Fig.~\ref{fig:comp1k}, our TMFNet  focuses better on the interesting objects than the baseline method and other SOTA methods \cite{gca,indexnet} with interference from salient background objects. More details are reported in supplemental materials.

\begin{figure*}[t]
	\centering
	\includegraphics[width = 0.94\textwidth]{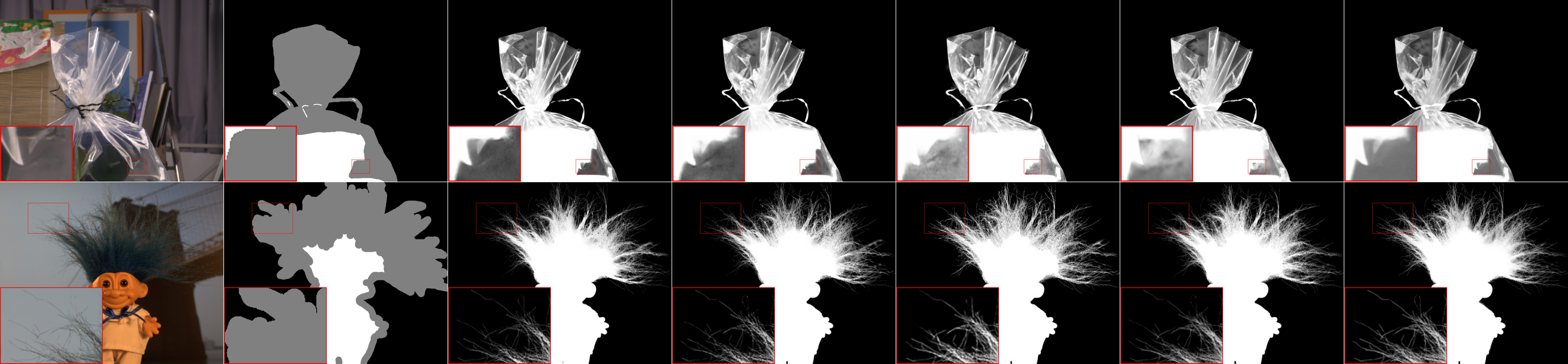}
	\caption{The visual comparison results on Alphamatting benchmark. From left to right, the original image, trimap, A$^{2}$U \cite{a2u}, GCA \cite{gca}, ADA \cite{adamatting}, SIM \cite{sim} and ours. }
	\label{fig:alpharesult}
\end{figure*}

\textbf{Alphamatting benchmark.} Compared with other state-of-the-art methods such as LFPNet \cite{lfpnet}, SIM \cite{sim}, ADA \cite{adamatting}, GCA \cite{gca} and A$^2$U \cite{a2u}, our method performances better on metrics of both SAD and MSE, shown in Table~\ref{tab:alphamatting}. Several visual results are shown in Fig.~\ref{fig:alpharesult} and our methods also have better visual performance on these real-world cases in alphamatting \cite{alphamatting} benchmark.

\textbf{High-resolution real-world images.} Besides real-world cases in Alphamatting \cite{alphamatting} benchmark, we also collect  several high-resolution real-world images and  draw trimaps for them. As is shown in Fig.~\ref{fig:real}, our model has a better prediction for details than the SOTA methods \cite{deepmatting,indexnet,gca} for high-resolution real-world testing cases.

\begin{figure*}[h]
	\centering
	\includegraphics[width = 0.94\textwidth]{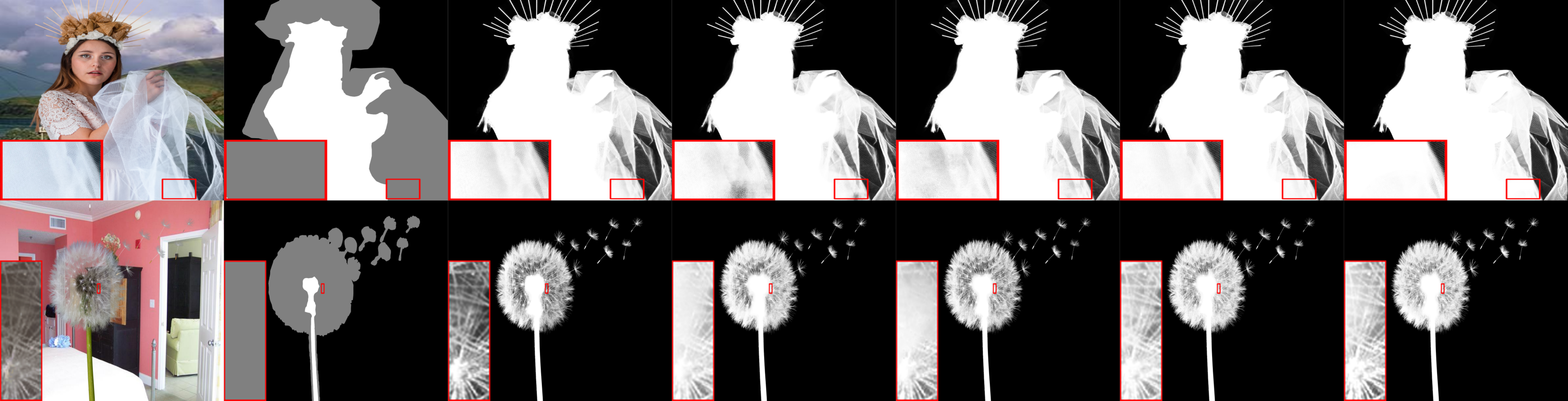}
	\caption{The visual comparison results on our CIOM test set. From left to right, the original image, trimap, IndexNet \cite{indexnet}, GCA \cite{gca}, baseline, ours and ground-truth. }
	\label{fig:ciom}
	\vspace{-2.2em}
\end{figure*}

\textbf{Our CIOM test set} provides quantitative results for high-resolution and high-quality images with a resolution up to 4000$\times$3867. As is shown in Table~\ref{tab:ciom}, our proposed TMP module and GLF module bring 4.6 and 2.2 SAD improvements, respectively. We also compare our method with DIM \cite{deepmatting}, IndexNet \cite{indexnet}, and  GCA \cite{gca} trained on our CIOM training set. All testings are implemented on a 32GB Tesla V100 GPU. Since GCA \cite{gca} can only evaluate images with a resolution up to 3000$^2$ with 32GB memory, we downsample images larger than 3000$^2$ for the testing of GCA. Our method achieves 20.2 SAD, which outperforms other methods significantly.  Visual comparison for proposed methods, baseline, and some of those methods can be seen in Fig.~\ref{fig:ciom}. Our method outperforms the above methods on both quantitative and visual results on this high-resolution matting benchmark.
\begin{table}[t]
    \begin{center}
    \caption{Ablations and comparison on the Composition-1k test set.}
    \begin{tabular}{lccc}
        \hline
        Methods & SAD & MSE   & Params \\
        \hline
        base loss:\\
        Basic & 28.1 & 5.8 & 34.8M \\
        Basic+TP &27.1 & 5.1 & 34.8M \\ 
        Basic+MP&27.9 & 5.4 & 34.8M\\
        Basic+TMP&26.9 & 5.1 & 34.8M\\
        Basic+TMP+LF&26.3&5.1&33.8M\\
        Basic+TMP+GLF&\textbf{24.9} & \textbf{4.8} & 33.9M\\
        Basic+TMP+GLF(B)&26.2 & 5.0 & 33.9M\\
        Basic+TMP+GLF(C)&26.7 & 5.0 & 34.1M\\
        \hline
        Comparison Methods\\
        base loss:\\
        Basic+ASPP\cite{v3}&27.9 & 5.3 & 41.2M\\
        Basic+TMP+CARAFE\cite{carafe}&28.6 & 5.7 & 35.6M\\
        \hline
        +laplacian loss:\\
        Basic & 27.4 & 5.2 & 34.8M \\
        Basic+TMP&26.0 & 4.7 & 34.8M\\
        Basic+TMP+GLF&\textbf{24.0} & \textbf{4.1} & 33.9M\\
        \hline
    \end{tabular}
    \label{tab:albation}
    \end{center}
\end{table}

 \begin{table}[t]\small
    \vspace{-2.2em}
    \centering
    \caption{Quantitative results of our method and several representative state-of-the-art methods on Alphamatting \cite{alphamatting} benchmark. ``S'', ``L'', ``U'' denote three trimap sizes and scores denote average rank across 8 test samples. ``O'' denotes the overall average rank across ``S'', ``L'' and ``U''.}
    \setlength{\tabcolsep}{2mm}{
    \begin{tabular}{l|cccc|c|c}
    \hline
    \multicolumn{1}{c|}{\multirow{2}{*}{Methods}} &
    \multicolumn{4}{c|}{\multirow{1}{*}{SAD}} &
    \multicolumn{1}{c|}{\multirow{1}{*}{MSE}} &
    \multicolumn{1}{c}{\multirow{1}{*}{Grad}} \\
    \cline{2-7}
    \multicolumn{0}{c|}{} &
    \multicolumn{1}{c}{\multirow{1}{*}{O}} &
    \multicolumn{1}{c}{\multirow{1}{*}{S}} &
    \multicolumn{1}{c}{\multirow{1}{*}{L}} &
    \multicolumn{1}{c|}{\multirow{1}{*}{U}} &
    \multicolumn{1}{c|}{\multirow{1}{*}{O}} &
    \multicolumn{1}{c}{\multirow{1}{*}{O}} \\
    \hline
ADA \cite{adamatting} & 12.1 & 10.9 & 11.1 & 14.4 & 12.8  & 12.3 \\
A$^{2}$U \cite{a2u} & 12.5 & 11.4 & 9.8 & 16.3 & 14.6  & 11.3 \\
GCA\cite{gca} & 13.7 & 14.4 & 11.5 & 15.3 & 14.5 & 12.8  \\
SIM\cite{sim} & 5.8 & 6.3 & 5 & 6  & 6.3 & 6.2\\
LFP\cite{lfpnet}&4.5&3.8&3.5&6.4&4.1&\textbf{2.8}\\
Ours & \textbf{3.3} & \textbf{2.3} & \textbf{2.9} & \textbf{4.6}  & \textbf{4} & 3.9\\ 
    \hline     
    \end{tabular}}
    \label{tab:alphamatting}
\end{table}

\subsection{Ablation and Comparison}
Our TMP module consists of trimap-guided non-background average pooling and multi-scale pooling kernels, which mines the high-level semantic context of interesting objects pointed out by trimap. We report the ablation study in Table~\ref{tab:albation}. The ``Basic+TP''  refers to replacing the adaptive average pooling with non-background adaptive average pooling in ``ppm'', which brings 1.0 SAD improvement by focusing the feature mining on non-background area without extra parameters. ``Basic+MP''  refers to replacing the adaptive average pooling with our multi-scale pooling with a stride of 1 whose smooth representation improves 0.2 SAD. Finally, ``Basic+TMP'' refers to replacing the ``ppm'' in baseline with our TMP module, which improves 1.2 and 1.4 SAD under base loss and with Laplacian loss, respectively. Besides ``ppm'' in baseline, we also compare our TMP with ASPP \cite{v3} module, namely ``Basic+ASPP'' in Table~\ref{tab:albation} and our TMP outperforms ASPP with fewer parameters.

The ablation study for our GLF module can be seen in Table~\ref{tab:albation}. The ``LF'' refers to  a local-aware fusion module, namely  a GLF module without using global context feature $G$. The ``Basic+TMP+LF'' improves 0.6 SAD from ``Basic+TMP'' and saves 1M parameters by replacing all the ``static fusion'' with our local-aware fusion modules. For selecting the global context feature for our GLF module, we compare GLF, GLF(B), and GLF(C) in Table~\ref{tab:albation}. The results show that GLF using global context from the output of our TMP module outperforms other selections significantly and it brings 1.4 SAD improvements with only 0.1M extra parameters. In total, ``Basic+TMP''+GLF improves 2.0 SAD and costs 0.9M fewer parameters. In addition, we also compare with the existing local-aware dynamic upsampling method such as CARAFE \cite{carafe}, which only uses high-level feature to predict upsampling kernels and shows a negative effect for matting task shown in Table~\ref{tab:albation}. Above analyses show that both global and local features are important for feature fusion in matting and our GLF module performs a global-local context-aware spatial fusion to improve natural image matting both efficiently  and effectively. Ablation studies of fusion stages are in Appendix~\ref{app:fusion_stage}.

\subsection{Visualization of Our Fusion Kernels}
We visualize our predicted fusion kernel maps of the initial network, a trained local-aware fusion module ``LF'', and a trained GLF module in Fig.~\ref{fig:kernel}. These kernel maps in the first row and the second row of Fig.~\ref{fig:kernel} are generated from the input cases of crystal and dandelion shown in Fig.~\ref{fig:comp1k}, respectively. Compared with the initial one, the trained local-aware fusion module  ``LF'' learns the structure of interesting objects, and  suppresses the interference of salient background objects to some degree. With the proper selection of a global context feature mined from our TMP module, our global-local context-aware fusion module (GLF) learns the structure of interesting objects better and its predicted fusion kernel maps have clearer boundaries. These visual results of predicted kernel maps show that the feature mining and the proper design of the fusion modules in our TMFNet are good for learning the structure of interesting objects in complex scenes.

\section{Conclusion}
In this paper, we observe that previous trimap-based matting methods lack an efficient way to integrate trimap guidance into semantic context feature mining for interesting objects and they also ignore the importance of a global context feature with high-level semantic information for feature fusion in matting. Based on this observation, we propose a trimap-guided feature mining and fusion network for  natural image matting. Our TMP module mines a powerful semantic context representation for interesting objects pointed out by trimap without extra parameters and our GLF uses the high-level semantic global context from TMP to promote our global-local context-aware feature fusion both efficiently and effectively. To advance the high-resolution and high-quality matting, we build a large-scale high-resolution dataset for common interesting object matting. Finally, extensive experiments demonstrate that our method outperforms the state-of-the-art methods.

\newpage
{\small\bibliographystyle{ieee_fullname}\bibliography{main}}

\clearpage
\newpage
\appendix
\section{Ablation for GLF Module in Fusion Stages}\label{app:fusion_stage}
As shown in Table~\ref{tabsupp:glf} for Composition-1k \cite{deepmatting} test set, based on ``Basic+TMP'', we gradually replace ``static fusion'' with local-aware fusion modules in the decoder at stages of F1, F2, and F3 in Figure~\ref{fig:fra}, and then the global context is gradually applied to these stages. When local-aware fusion modules are applied to all 3 fusion stages, it improves the SAD from 26.9 to 26.3 and saves about 1M parameters. And when the global context is applied to  all 3 fusion stages, it improves the SAD from 26.3 to 24.9 with only about 0.1M extra parameters. In total, our GLF modules  improve the SAD from 26.9 to 24.9 and save about 0.9M parameters.
\begin{table}[htb]
    \begin{center}
    \caption{Ablation for GLF module in fusion stages.}
    \begin{tabular}{lccc}
        \hline
        Fusion & SAD & MSE   & Params \\
        \hline
        base loss:\\
        Basic+TMP & 26.9 & 5.1 & 34.8M \\
        \hline
        base loss:\\
        Local feature only:\\
        F1&26.6 & 5.1 & 33.9M\\
        F1+F2 &26.4 & 5.1 & 33.8M \\ 
        F1+F2+F3&26.3 & 5.1 & 33.8M\\
        \hline
        base loss:\\
        +Global context:\\
        F1&26.2 & 5.0 & 33.8M\\
        F1+F2 &25.6 & 5.0 & 33.9M \\ 
        F1+F2+F3&\textbf{24.9} & \textbf{4.8} & 33.9M\\
        \hline
    \end{tabular}\label{tabsupp:glf}
    \end{center}
\end{table}
\section{Computation Costs}
We compare the computation costs of our TMFNet, the baseline model, GCA \cite{gca}, and FBA \cite{fba}  under an input with a resolution of $2048\times 2048$ in Table~\ref{tabsupp:GFLOPs}. The proposed TMFNet  has lower computation costs than the baseline model and several SOTA methods \cite{gca,fba}. The results of GFLOPs are based on the way of calculation in the MMCV  \cite{mmcv} including every major operation in each model.
\begin{table}[h]
    \begin{center}
    \caption{Computation costs of models.}
    \begin{tabular}{l|c|c}
        \hline
        Methods & GFLOPs&Parameters  \\
        \hline
        GCA  \cite{gca} & 5385&25.3M \\
        FBA  \cite{fba} & 2741&34.7M \\
        \hline
        Baseline & 1410&34.8M \\
        Ours & 1121&33.9M\\
        \hline
    \end{tabular}

    \label{tabsupp:GFLOPs}
    \end{center}
\end{table}
\end{document}